\documentclass[11pt]{article}
\usepackage[final]{acl}

\usepackage{microtype}
\usepackage{amsmath}
\usepackage{wrapfig} 
\usepackage{amssymb}
\usepackage{graphicx}
\usepackage{mdframed}
\usepackage{booktabs}
\usepackage{times}
\usepackage[table,xcdraw]{xcolor}
\usepackage{multirow} 
\usepackage{hyperref}
\usepackage{float}
\usepackage{xcolor}
\usepackage{listings}
\definecolor{prompt_background}{HTML}{f4faed}
\usepackage[english,bidi=default]{babel} 
\babelprovide[import]{hindi}
\babelprovide[import]{arabic}

\title{Aligning Paralinguistic Understanding and Generation in Speech LLMs\\via Multi-Task Reinforcement Learning}

\author{\normalsize
    \textbf{Minseok Kim\thanks{equal contribution. Correspondence to: \{kminseok, seanchen, zhaojiang\}@meta.com}}, 
    \textbf{Jingxiang Chen\footnotemark[1]}, 
    \textbf{Seong-Gyun Leem}, 
    \textbf{Yin Huang}, 
    \textbf{Rashi Rungta}, 
    \\
\normalsize
    \textbf{Zhicheng Ouyang}, 
    \textbf{Haibin Wu}, 
    \textbf{Surya Teja Appini}, 
    \textbf{Ankur Bansal}, 
    \textbf{Yang Bai}, 
    \\
\normalsize
    \textbf{Yue Liu}, 
    \textbf{Florian Metze}, 
    \textbf{Ahmed A Aly}, 
    \textbf{Anuj Kumar}, 
    \textbf{Ariya Rastrow}, 
    \textbf{Zhaojiang Lin\footnotemark[1]} 
    \\
    \\
    Meta Reality Labs
}

\begin{document}

\maketitle
\begin{abstract}
Speech large language models\,(LLMs) observe paralinguistic cues such as prosody, emotion, and non-verbal sounds—crucial for intent understanding. However, leveraging these cues faces challenges: limited training data, annotation difficulty, and models exploiting lexical shortcuts over paralinguistic signals. We propose multi-task reinforcement learning\,(RL) with chain-of-thought prompting that elicits explicit affective reasoning. To address data scarcity, we introduce a paralinguistics-aware speech LLM\,({\sc PALLM}) that jointly optimizes sentiment classification from audio and paralinguistics-aware response generation via a two-stage pipeline. Experiments demonstrate that our approach improves paralinguistics understanding over both supervised baselines and strong proprietary models\,(Gemini-2.5-Pro, GPT-4o-audio), by 8-12\% on Expresso, IEMOCAP, and RAVDESS. The results show that modeling paralinguistic reasoning with multi-task RL is crucial for building emotionally intelligent speech LLMs. 
\end{abstract}

\section{Introduction}
Spoken interaction is becoming a primary interface for large language models (LLMs), driven by recent speech LLMs that accept speech as input and produce natural-language responses~\cite{zeng2024glm4voice, xu2025qwen25omni, huang2025stepaudio, wu2024towards,arora2025landscape}.
Unlike text-only models, speech LLMs have access to not only lexical content but also paralinguistic cues such as prosody, emotion, speaking style, and non-verbal sounds from a user's input. These cues are often decisive for determining communicative intent: the same utterance (e.g., ``I got 80\% on my test'') may call for celebration when delivered in a cheerful tone or comfort when expressed with disappointment. Systems that respond only to transcripts risk being semantically correct yet emotionally misaligned, undermining user trust and perceived empathy.
While speech LLMs' access to paralinguistic information presents significant opportunities, effectively leveraging this information for contextually appropriate conversational behavior remains challenging. 

Recent work has explored paralinguistic processing in speech LLMs through two primary approaches: (i) speech emotion recognition (SER)~\cite{li2025emorl}, treating emotion detection as a classification task, and (ii) paralinguistics-aware response generation~\cite{wu2025stepaudio2}, focusing on curating large-scale audioset for supervised fine-tuning (SFT). However, a fundamental challenge in developing paralinguistics-aware generation systems is the scarcity of suitable training data and the difficulty of annotating ground truth for emotionally appropriate responses. Unlike emotion classification, which can rely on established taxonomies, determining whether a response exhibits appropriate emotional alignment requires nuanced human judgment that is both subjective and context-dependent.

Furthermore, SFT alone faces inherent limitations in learning robust paralinguistic awareness. When textual content already suggests sentiment (e.g., ``I failed my exam''), models can minimize training loss by relying on lexical cues while bypassing prosodic information, potentially yielding responses that appear plausible but remain insensitive to subtle tonal variations or non-verbal cues such as sighs or laughter. This challenge is compounded when lexical and paralinguistic cues conflict (e.g., ``I'm fine'' spoken with distressed prosody), highlighting the necessity for models to be explicitly grounded in both the understanding and generation of paralinguistic information.

\begin{figure}[t]
\vspace{-0.5cm}
    \centering
    \includegraphics[width=0.5\textwidth]{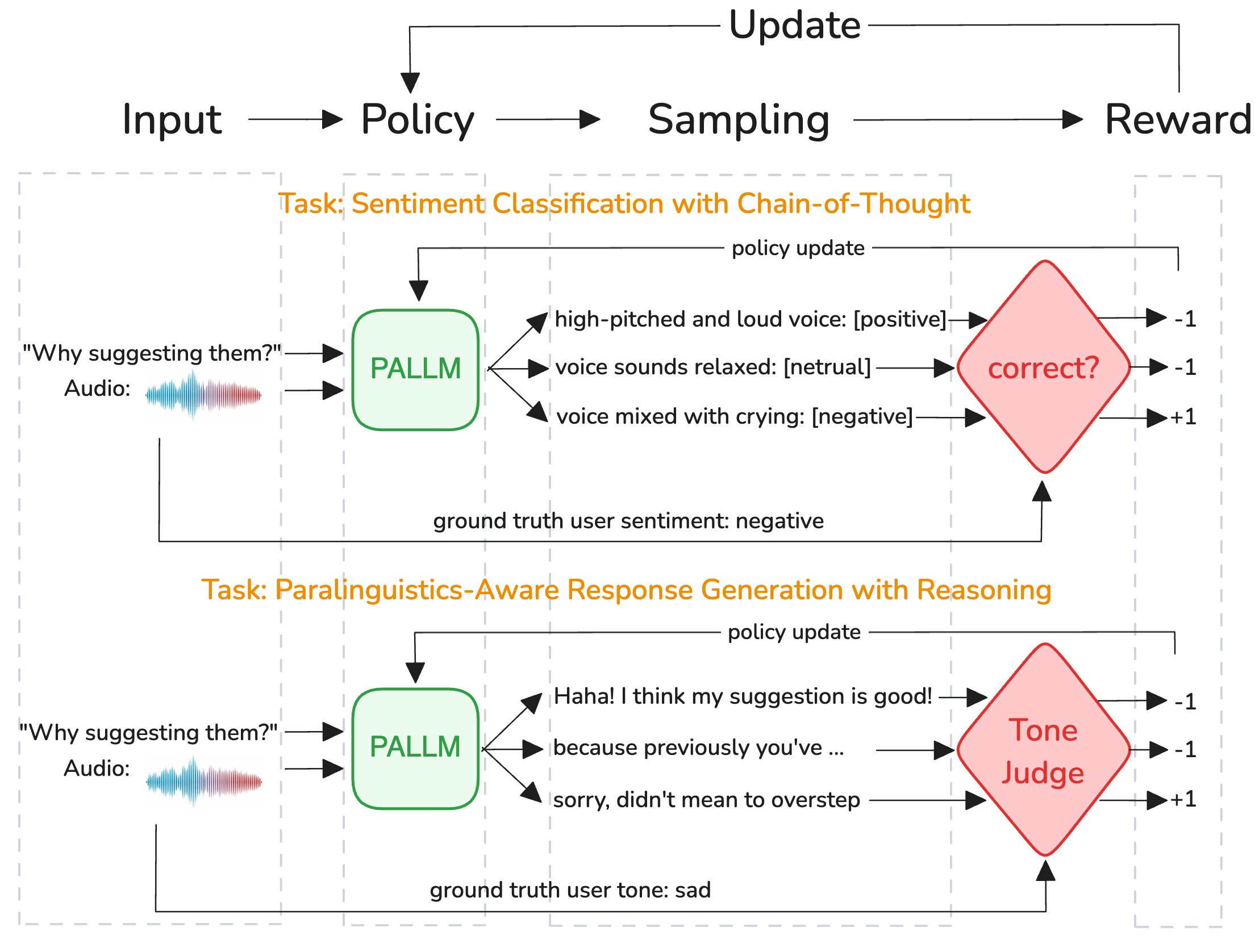}
    \caption{Paralinguistics-Aware LLM stage 2 overview. A multi-task RL jointly performs sentiment classification and paralinguistics-aware response generation with chain-of-thought reasoning.}
    \label{fig:tone_ARPG}
\vspace{-0.5cm}
\end{figure}

In this work, we address these challenges by proposing a Paralinguistics-Aware LLM ({\sc PALLM}) that jointly learns (i) sentiment classification of the user's spoken utterance and (ii) response generation whose style aligns with the inferred affect. Our training procedure consists of two stages: In \textbf{Stage 1}, we perform supervised fine-tuning on sentiment labels and synthesized paralinguistics-aware responses to establish the model’s foundational ability to recognize and generate responses sensitive to paralinguistic cues.
In \textbf{Stage 2}, we apply online reinforcement learning on two coupled tasks: sentiment classification with chain-of-thought\,(CoT) reasoning and paralinguistics-aware response generation.
This stage further enhances the model’s paralinguistic understanding by explicitly grounding both sentiment classification and response generation in audio-based evidence through RL with CoT reasoning.
The training paradigm for Stage 2 is illustrated in Figure~\ref{fig:tone_ARPG}.


Our main contributions are as follows: \textbf{First}, we formalize paralinguistic awareness in speech LLMs as a multi-task RL reasoning problem. It jointly learns \emph{(i) sentiment classification} from acoustic–prosodic cues and \emph{(ii) paralinguistics-aware response generation}. 
\textbf{Second}, we propose {\sc PALLM}, a two-stage training pipeline that first performs joint SFT on sentiment labels and synthesized tone-conditioned responses, and then applies multi-task RL to reduce reliance on lexical shortcuts and explicitly ground decisions in paralinguistic evidence from audio. 
\textbf{Third}, we conduct a comprehensive evaluation on Expresso, IEMOCAP, and RAVDESS, comparing against SFT-only baselines and strong proprietary speech LLMs (Gemini-2.5 Pro, GPT-4o-audio). The results show that {\sc PALLM} consistently improves the response appropriateness significantly and hence shows a better paralinguistics understanding, supported by automatic and human evaluations.


\section{Related Work}

\subsection{Speech Emotion Recognition and Paralinguistic Modeling}

SER has traditionally been formulated as a classification task using hand-crafted acoustic features or deep learning on spectrograms~\cite{el2011survey,schuller2013interspeech}.
Recent work leverages self-supervised speech representations from models such as emotion2vec~\cite{ma2023emotion2vec} and HuBERT~\cite{hsu2021hubert}, achieving strong results on benchmarks like IEMOCAP~\cite{busso2008iemocap, wagner2023dawn}.

With the emergence of large language models, several approaches integrate SER into LLM-based frameworks.
AA-SLLM and SECap use external audio encoders to extract emotion features and bridge them to frozen LLMs for emotion classification or captioning~\cite{mai2025aasllm,xu2024secap,liang2024aligning}.
More recent audio-language models such as EMO-RL formulate SER as a generative reasoning problem with CoT prompting, applying GRPO-style\,\cite{shao2024deepseekmath} RL to improve emotional reasoning~\cite{li2025emorl}.
While these methods achieve strong classification performance, they primarily target SER as an isolated task without mechanisms to translate detected affect into conversational responses.

\subsection{Paralinguistic-Aware Dialogue Systems}

Several recent works extend spoken dialogue systems to incorporate paralinguistic information.
In text-based settings, empathetic dialogue systems jointly model emotion recognition and response generation, demonstrating that understanding affect improves response quality~\cite{rashkin2019towards,majumder2020mime}.
For speech-based interaction, ParalinGPT conditions LLMs on speech embeddings and sentiment attributes for multi-task prediction~\cite{lin2024paralinguistics}, while E-chat and EMOVA integrate emotion representations into LLMs for affective conversation~\cite{xue2024echat,chen2025emova}.

Speech LLMs such as GLM-4-Voice~\cite{zeng2024glm4voice}, Qwen2-Audio~\cite{chu2024qwen2audio}, and Step-Audio~2~\cite{huang2025stepaudio} process speech inputs directly for emotion-aware capabilities.
Concurrently, ParaS2S introduces a benchmark and GRPO-based framework for paralinguistic-aware speech-to-speech dialogue~\cite{yang2025paras2s}, while Step-Audio~2 applies ``reasoning-centric'' RL for expressive audio interaction~\cite{wu2025stepaudio2}.

These systems illustrate growing interest in paralinguistic dialogue, yet they face fundamental limitations in how paralinguistic awareness is achieved. Most previous works rely on external emotion encoders or focus on speech-to-speech generation, while critically, they either optimize SER in isolation or train generation models without explicit emotion understanding objectives. This decoupling creates vulnerability to lexical shortcuts, where models infer user emotions primarily from textual content rather than acoustic-prosodic cues.
To our knowledge, no prior work jointly optimizes sentiment classification and paralinguistics-aware generation through multi-task RL with CoT-structured reasoning for speech LLMs. Our approach addresses this by requiring explicit reasoning about paralinguistic evidence, enabling mutual reinforcement between affect perception and appropriate response generation.




\section{Methodology}
We frame paralinguistic awareness as a multi-task problem where a speech LLM must jointly (1) classify the sentiment of a spoken utterance from acoustic-prosodic cues, and (2) generate responses whose emotional tone is appropriate given the inferred affect.
We train the LLM in two stages: SFT to cold-start a paralinguistic-aware speech LLM base policy, followed by RL with CoT reasoning to refine both understanding and generation capabilities for paralinguistics.
\subsection{Task Formulation}
\subsubsection{Sentiment Classification}
Given a spoken utterance represented as audio $\mathbf{a}$, the model predicts a sentiment label $s \in \{\text{positive, neutral, negative}\}$ by interpreting the user's emotional state from acoustic and prosodic cues.
We choose coarse-grained sentiment categories over fine-grained tone labels (e.g., happy, sad, angry, fearful) for two practical reasons.
First, fine-grained tone taxonomies vary across datasets and application domains, limiting cross-dataset generalization.
Second, semantically similar tones (e.g., ``happy'' vs.\ ``cheerful,'' ``depressed'' vs.\ ``sad'') are difficult for models to distinguish, and conflating them during training can confuse the model.
Coarse sentiment categories provide a more stable and generalizable representation of user affect while retaining sufficient granularity for contextually appropriate response generation.
\subsubsection{Paralinguistics-Aware Response Generation}
Given the same audio input $\mathbf{a}$, the model generates a textual response $\mathbf{r}$ whose emotional tone is coherent with the user's current affective state.
For example, the utterance ``I got 80\% on my test'' requires an empathetic, comforting response when spoken with a sad tone, but a celebratory response when spoken cheerfully.
This task refines the model's ability to translate paralinguistic understanding into contextually appropriate conversational behavior, moving beyond semantically correct but emotionally tone-deaf responses.
\subsection{Two-Stage Training Pipeline}
\subsubsection{Stage 1: Supervised Fine-Tuning}
We initialize the model with joint SFT on sentiment classification and paralinguistics-aware response generation.
This stage is essential because paralinguistic cues are highly sparse in typical conversational data, making RL-only approaches ineffective without a warm start.
\paragraph{(SFT) Sentiment Classification}
Given audio input $\mathbf{a}$ and ground-truth sentiment $s$ converted from fine-grained tone annotations $l$ using rule-based mapping\,(e.g., ``happy'' label to ``positive'' label, see Appendix for label mapping details), we minimize cross-entropy loss:
$$\mathcal{L}_{\text{cls}} = -\log P(s \mid \mathbf{a}; \theta)$$
This task provides explicit supervision for affect detection, encouraging the model to attend to acoustic-prosodic features.
\paragraph{(SFT) Paralinguistics-Aware Response Generation}
Since our training data lacks ground-truth emotionally appropriate responses, we synthesize them by prompting an external text LLM to generate responses conditioned on the transcript $\mathbf{t}$, which is the ASR output of audio input $\mathbf{a}$, and ground-truth tone annotation $l$.
While these synthesized responses lack access to fine-grained paralinguistic details from audio (e.g., hesitations, laughter, sighs), they provide a useful initialization for tone-conditioned generation.
We minimize the following generation loss:
$$\mathcal{L}_{\text{gen}} = -\sum_{i=1}^{|\mathbf{r}^*|} \log P(r^*_i \mid r^*_{<i}, \mathbf{a}, \mathbf{t}; \theta)$$
where $\mathbf{r}^*$ is the synthesized response.
We jointly optimize both tasks with equal weighting:
$$\mathcal{L}_{\text{SFT}} = \mathcal{L}_{\text{cls}} + \mathcal{L}_{\text{gen}}$$
\subsubsection{Stage 2: Reinforcement Learning with Chain-of-Thought}
SFT alone has two critical limitations.
First, when textual content already hints at sentiment (e.g., ``I failed my exam''), models can minimize training loss by relying on lexical-semantic correlations while bypassing acoustic-prosodic processing.
Second, responses synthesized by text LLMs cannot capture subtle paralinguistic nuances that distinguish genuinely empathetic interactions from generic, emotionally superficial ones.
To address these limitations, we introduce a reinforcement learning stage with explicit CoT reasoning.
As illustrated in Figure~\ref{fig:tone_ARPG}, our approach requires the model to articulate \emph{why} it classifies an utterance with a particular sentiment and \emph{how} that sentiment informs its response strategy before producing final outputs.
This explicit reasoning mechanism discourages lexical shortcuts by forcing the model to ground its predictions in paralinguistic evidence from audio.

\paragraph{(RL) Sentiment Classification with CoT}
The policy model receives audio input $\mathbf{a}$ and generates a reasoning trace $\mathbf{c}$ followed by a sentiment prediction $\hat{s}$:
$$\pi_\theta(\mathbf{a}) \rightarrow \langle \mathbf{c}, \hat{s} \rangle$$
We use a rule-based judge to verify correctness, yielding a binary reward:
$$r_{\text{cls}} = \mathbb{1}[\hat{s} = s]$$
where $r_{\text{cls}} \in \{-1, 1\}$. This forces the model to ground its predictions in paralinguistic evidence rather than lexical shortcuts.
For example, a reasoning trace might state: \textit{``The speaker's hesitant prosody, prolonged pauses, and low pitch contour suggest negative sentiment, despite neutral lexical content.''} before \textit{``negative''} sentiment prediction. 
\paragraph{(RL) Paralinguistics-Aware Response Generation with Reasoning}
Similarly, the model generates reasoning $\mathbf{c}'$ about the user's affective state, followed by a response $\hat{\mathbf{r}}$:
$$\pi_\theta(\mathbf{a}) \rightarrow \langle \mathbf{c}', \hat{\mathbf{r}} \rangle$$
We employ an LLM judge~\footnote{Details of judge models and prompts are available in Appendix~\ref{apdx_a.1}} to evaluate whether $\hat{\mathbf{r}}$ exhibits appropriate emotional tone given the transcript $\mathbf{t}$ and ground-truth emotion.
The LLM judge evaluates responses against a criteria rubric and outputs binary labels, which are then converted to binary scores: $r_{\text{gen}} \in \{-1, 1\}$. 

\paragraph{Policy Optimization}
We optimize the policy model via GRPO~\cite{shao2024deepseekmath} to maximize expected advantage using group-relative returns. To enable multi-task learning, we construct separate prompts for CoT classification and paralinguistics-aware generation tasks, and apply task-specific rewards $r_{\text{cls}}$ and $r_{\text{gen}}$ respectively. The model parameters are updated via policy gradients.


\section{Experiments}
\subsection{Datasets}
We evaluated the paralinguistics-awareness of models on three datasets: Expresso\,\cite{nguyen2023expresso}, IEMOCAP\,\cite{busso2008iemocap}, and RAVDESS\,\cite{livingstone2018ravdess}.
To ensure relevance to conversational scenarios, we filtered out examples with fewer than 1 word or more than 20 words across all datasets.
\begin{wraptable}{r}{0.5\columnwidth} 
\vspace{-\baselineskip}              
\centering
\small
\resizebox{\linewidth}{!}{
\begin{tabular}{crr}
\toprule
\multicolumn{1}{c}{dataset} & \multicolumn{1}{c}{train} & \multicolumn{1}{c}{eval} \\
\midrule
Expresso             & 12,878                     & 3,031                     \\
IEMOCAP              & 6,738                      & 844                      \\
RAVDESS              & N/A                         & 1,248                    \\
\bottomrule
\end{tabular}
}
\caption{Dataset statistics.}
\label{tab:dataset_stats}
\vspace{-1\baselineskip}
\end{wraptable}
For Expresso, we perform speaker-level splits by randomly selecting two speakers as the held-out test set and using the remaining speakers for training, preventing speaker identity leakage.
For IEMOCAP, we randomly sample 10\% of utterances for evaluation and use the remaining 90\% for training.
RAVDESS is held out entirely from training to assess out-of-distribution generalization to unseen paralinguistic data.
Table~\ref{tab:dataset_stats} summarizes the resulting statistics for the three datasets.

\begin{table*}[t]
\centering
\small
\begin{tabular}{c|ccc|ccc}
\hline
\rule{0pt}{2.5ex}                                      & \multicolumn{3}{c|}{\textbf{Sentiment Classification}}                                               & \multicolumn{3}{c}{\textbf{Response Appropriateness}} \\ [2pt] \cline{2-7} 
\rule{0pt}{2.5ex} \multirow{-2}{*}{\textbf{Model Name}} & \textbf{Expresso} & \textbf{IEMOCAP} & \textbf{RAVDESS} & \textbf{Expresso} & \textbf{IEMOCAP} & \textbf{RAVDESS} \\[2pt] \hline
\rule{0pt}{2.5ex} Gemma-3n                              & 39.7\%                        & 48.1\%                        & 23.2\%                        & 59.0\%           & 57.5\%          & 30.2\%          \\
Qwen-2.5                              & 42.4\%                        & 38.0\%                        & 24.4\%                        & 59.6\%           & 55.7\%          & 36.9\%          \\
Gemini-2.5 Flash                      & 47.0\%                        & 52.8\%                        & \textbf{61.4}\%                        & 51.6\%           & 41.0\%          & 31.3\%          \\
Gemini-2.5 Pro                        & 53.7\%                        & 54.0\%                        & 44.2\%                        & 66.1\%           & 57.2\%          & 37.7\%          \\
GPT4o-Audio                           & 39.9\%                        & 46.2\%                        & 28.3\%                        & 67.4\%           & 61.4\%          & 39.7\%          \\ \hline
\rule{0pt}{2.5ex} {\sc SFT (Gen only)}                      & 41.0\%                        & 46.0\%                        & 28.0\%                        & 61.0\%           & 57.0\%          & 30.0\%          \\
{\sc SFT (Cls + Gen)}                       & \textbf{74.0\%}                        & \textbf{59.0\%}                        & 54.0\%                        & 65.0\%           & 59.0\%          & 36.0\%          \\ \hline
\rule{0pt}{2.5ex} {\sc PALLM (Gen only)}              & \textbf{74.0\%}                        & 56.0\%                        & 57.0\%                        & 73.0\%           & 70.0\%          & 44.0\%          \\
{\sc PALLM (Cls + Gen)}           & \textbf{74.0\%}                        & 57.0\%                        & 59.0\%                        & \textbf{77.0\%}           & \textbf{73.0\%}          & \textbf{48.0\%}          \\ \hline
\end{tabular}
\caption{Performance comparison on Expresso, IEMOCAP, and RAVDESS datasets, evaluating sentiment accuracy and response appropriateness. \textbf{Bold} font indicates best performance among all models. Our multi-task RL approach {\sc PALLM (Cls + Gen)} consistently achieves the best response appropriateness across all datasets while maintaining competitive sentiment classification accuracy.}
\label{tab:main_results}
\end{table*}

\subsection{Implementation}
We employed the Llama 4 Scout (17Bx16E)\footnote{\url{https://www.llama.com/}} model as the foundational backbone for our experiments, with additional speech understanding capabilities integrated as described in Llama 3 speech paper \cite{dubey2024llama}. We train our LLM parameters with audio encoder frozen. For multi-task RL, we sample the CoT classification and paralinguistic generation tasks uniformly, and for each training batch, we perform $K=4$ generations, compute advantages using group-relative returns, and update parameters via policy gradients.

We benchmark PALLM against state-of-the-art approaches. We name {\sc SFT (Gen only)} that performs paralinguistics-aware response generation of SFT following \cite{zhou2018emotional}, and {\sc SFT (Cls + Gen)} which performs both SFT tasks following \cite{ide2021multi}.
Note that the baseline {\sc SFT (Cls + Gen)} has been used as a pickup checkpoint for our RL models, namely {\sc PALLM (Gen only)} that is only trained with paralinguistics-aware response generation with reasoning RL task and {\sc PALLM (cls + Gen)} that is trained with both RL tasks.
\ref{appendix:training_prompt} shows the instruction prompts used for SFT and RL stages.

We also evaluate popular speech models, including both open-source speech LLMs (Gemma-3n\,\cite{gemma3_tech_report_2025}, Qwen-2.5\,\cite{ qwen25_tech_report_2025}) and proprietary speech LLMs (Gemini-2.5 Flash, Gemini-2.5 Pro\,\cite{gemini25_report_2025}, GPT-4o Audio\,\cite{hurst2024gpt}). 
We exclude SER-only models (e.g., \cite{wagner2023dawn}) from benchmarking because they are not speech-capable LLMs and thus cannot generate responses.

\subsection{Metrics}
We assess two primary capabilities:\emph{(1) sentiment classification accuracy}, computed by binning tone predictions into positive, neutral, and negative categories and comparing with ground truth, and \emph{(2) paralinguistics-aware response appropriateness}, judged by LLM-based evaluator using GPT-4o with ground truth user tone whose evaluation prompt is in Appendix\,\ref{apdx_a.1}.
We also conducted a human evaluation using the same guideline prompt for unbiased assessment, where we found 82\% agreement with the GPT-4o judge.

\subsection{Results}

\subsubsection{Overall Results}
Table\,\ref{tab:main_results} presents a comparison of the proposed method {\sc PALLM(Cls + Gen)} with the baselines.
As shown in the table, our proposed approach showcases substantial improvements across benchmarks, especially in response appropriateness in all datasets.
Notably, our approach outperforms strong proprietary baselines including Gemini-2.5 Pro (66.1\% vs. 77.0\% appropriateness on Expresso) and GPT-4o-Audio (67.4\% vs. 77.0\%).

The progression from {\sc SFT (Cls + Gen)} to {\sc PALLM (Gen only)}, then to {\sc PALLM (Cls + Gen)} demonstrates the effectiveness of our two-stage training pipeline.
For Expresso as an example: we observe that 65.0\% of {\sc SFT (Cls + Gen)} is improved to 73.0\% in {\sc PALLM (Gen only)} if we perform RL response generation task, and adding classification further boosts performance to 77.0\%. 
This confirms that explicit sentiment grounding prevents lexical shortcuts and enables genuine understanding of paralinguistics during model training.
In other words, training of classification and generation proves the synergy between the two tasks is crucial.

\begin{table*}[!htbp]
\centering
\scriptsize
\setlength{\tabcolsep}{3pt}
\begin{tabular}{p{0.13\textwidth} p{0.06\textwidth} p{0.41\textwidth} p{0.35\textwidth}}
\toprule
\textbf{Utterance} & \textbf{Tone} & \textbf{SFT (Cls + Gen)} & \textbf{PALLM (Cls + Gen)} \\
\midrule
I understand. Okay. & Fearful & It seems like you're processing some information. Is there something specific you'd like to talk about? & Hey, it's okay. Take a deep breath. Want to talk about what's on your mind? \\
\midrule
Okay, I've never done this before. & Fearful & Don't worry, you're all set! We'll figure it out together. & It's totally normal to feel nervous. We'll figure it out together. \\
\midrule
I did. & Sad & It seems like you started to share something. Would you like to continue? & Sorry to hear that didn't go as planned. Want help figuring out what went wrong? \\
\midrule
Okay & Laugh & So, what's on your mind? & Want me to play some music or we can chat? \\
\midrule
Let's make a pact? & Cheerful & Sounds like a plan. What's the pact about? & Ah, I'm in, what's the pact? Want me to help you come up with something? \\
\bottomrule
\end{tabular}
\caption{Response generation examples of {\sc SFT (Cls + Gen)} vs {\sc PALLM (Cls + Gen)}.}
\label{tab:generation_examples}
\end{table*}

\subsubsection{Evaluation via Human Review}
To complement our automatic evaluation, we conducted human assessment on 100 random Expresso examples. 
We generated responses using GPT-4o Audio, {\sc SFT (Cls + Gen)} and {\sc PALLM (Cls + Gen)}, then asked annotators to blindly judge whether the response demonstrates good tone understanding. 
The models achieved 68\%, 62\%, and 76\% appropriateness, consistent with the improvements in automatic evaluation.

\subsubsection{Qualitative Analysis}

We selected representative examples where user tone and sentiment are ambiguous in text-only format but clear in audio, demonstrating our best model's performance on both paralinguistic-aware generation and classification tasks.
Table~\ref{tab:generation_examples} illustrates paralinguistic-aware response generation comparing {\sc SFT (Cls + Gen)} and {\sc PALLM (Cls + Gen)}. 
The improvements demonstrate how multi-task RL training teaches the model to translate affect perception into contextually appropriate responses. 
For fearful utterances like ``I understand. Okay'', {\sc SFT (Cls + Gen)}  produces neutral, generic responses, while {\sc PALLM (Cls + Gen)} offers emotional support and calming language. 
For the sad utterance ``I did.'', the model shifts from vague continuation prompts to empathetic problem-solving. The model also learns to match playful energy for laughing speech and enthusiastic tone for cheerful utterances. 
Critically, none of these utterances contain explicit emotion words—their emotional meaning derives entirely from prosodic delivery, demonstrating that our approach develops genuine paralinguistic processing rather than exploiting lexical shortcuts.

\section{Conclusion}

This work demonstrates that explicit paralinguistic reasoning through multi-task SFT and RL training significantly improves speech LLMs' ability to understand and respond to user affect. 
Our approach achieves substantial gains over proprietary baselines including GPT-4o Audio, with response appropriateness improving from 67.4\% to 77.0\% on Expresso. 
Joint training of sentiment classification and paralinguistics-aware generation proves essential: explicit sentiment grounding prevents lexical shortcuts and enables genuine paralinguistic awareness in speech LLMs.

\section{Limitations}
While our proposed approach achieves significant improvements in paralinguistic awareness, we observe several limitations.
First, we see a gap between the performance on in-domain (Expresso and IEMOCAP) and out-of-domain (RAVDESS) datasets, highlighting the need to address domain shift and improve coverage.
Second, our reliance on emotion labels in training datasets, which are required for both sentiment classification and paralinguistics-aware response generation in the RL stage, potentially limits the ability to leverage unlabeled audio datasets during training, which could improve coverage.
Third, the use of LLM-as-a-judge as a reward model for paralinguistics-aware response generation in the RL stage is constrained by challenges such as potential bias in the judge and vulnerability to reward hacking.

\bibliography{custom}

\clearpage
\appendix

\section{Appendix}
\label{sec:appendix}

\subsection{Tone to Sentiment Mapping}
Table~\ref{tab:tone_sentiment_mapping} presents the mapping of tone labels to sentiment categories for each dataset. Audio samples labeled as `surprised' were excluded from our analysis, as they can correspond to both positive and negative contexts, making it challenging to reliably distinguish their sentiment without significant additional effort.

\begin{table*}[ht]
\centering
\begin{tabular}{|l|l|l|l|}
\hline
\textbf{Sentiment} & \textbf{Expresso} & \textbf{IEMOCAP} & \textbf{RAVDESS} \\
\hline
positive & laughing, happy & excited, happy & happy \\
\hline
neutral & neutral & neutral & neutral, calm \\
\hline
negative & angry, sad, fearful & angry, sad, fear, frustrated, disgust & sad, angry, fearful, disgust \\
\hline
\end{tabular}
\caption{Mapping of tone to sentiment in Expresso, IEMOCAP, and RAVDESS datasets.}
\label{tab:tone_sentiment_mapping}
\end{table*}

\subsection{Response Appropriateness Instruction Prompt}
\label{apdx_a.1}

To evaluate paralinguistic-aware response appropriateness, we develop an LLM-as-a-judge that consumes \textbf{conversation history}, \textbf{user utterance with ground-truth tone}, and \textbf{assistant response}. It outputs a binary decision (YES/NO). The detailed prompt is shown in Figure \ref{fig:response_appropriateness_p1} and \ref{fig:response_appropriateness_p2}:

\begin{figure*}[h!]
\begin{mdframed}[backgroundcolor=prompt_background,linewidth=0.0pt]
\texttt{\tiny 
\#\# [Task]\\
You are an LLM tasked with judging whether an AI assistant's response content appropriately matches the user's tone in a multi-turn or single-turn conversation.\\
\\
\#\# [Persona guidelines]\\
\\
The assistant's persona is: a friendly AI assistant designed specifically for natural, conversational interactions.\\
\\
The assistant should respond in a way that:\\
* Appropriately acknowledges and validates the user's emotional state when necessary\\
* Maintains a friendly, conversational tone consistent with the persona\\
* Adjusts language and phrasing to match the emotional context of the conversation\\
\\
You will be given the conversation history, last user turn (along with a tone tag in brackets), and assistant's response.\\
\\
\#\# Your Task: Evaluate Tone Matching\\
\\
Assess whether the assistant's **response content** appropriately matches the user's emotional state and intent.\\
\\
**Look for:**\\
* Does the response content (words, phrasing) validate or appropriately respond to the user's emotional state?\\
* Is the language choice and tone appropriate for the user's emotional context?\\
* Does the response acknowledge the user's feelings when warranted?\\
* Is the response friendly, consistent with the persona?\\
\\
**Inappropriate responses include:**\\
* Mismatched emotions (e.g., cheerful when user is sad, dismissive when user is upset)\\
* Content that dismisses user feelings or contradicts their emotional state\\
* Responses that are cold, distant, or inconsistent with the friendly persona\\
* Overly emotional or dramatic responses to neutral queries\\
* Failure to acknowledge clear emotional states when acknowledgment is appropriate\\
\\
**Note:** Most neutral user queries should receive straightforward, friendly responses without excessive emotional framing.\\
\#\# [Step-by-Step Reasoning Instructions]\\
\\
1. **Identify the user's tone and emotional state** from their message\\
2. **Determine the appropriate response approach** - does the situation warrant emotional acknowledgment, or is a neutral/factual response more suitable?\\
3. **Examine the response content** - does the wording, phrasing, and overall tone match the user's state?\\
4. **Check consistency** - is the response friendly and conversational while being emotionally appropriate? If the user is seeking for factual information without emphasizing emotion in text, it is okay for the responses to be in neutral tone.\\
5. **Decide:**\\
   - If the response content appropriately matches -> Output ``YES''\\
   - If not -> Output ``NO'' with explanation\\
}
\end{mdframed}
\caption{Response appropriateness instruction prompt.}
\label{fig:response_appropriateness_p1}
\end{figure*}

\begin{figure*}[h!]
\begin{mdframed}[backgroundcolor=prompt_background,linewidth=0.0pt]
\texttt{\tiny 
**Judge output format:**\\
\{\{\\
  ``Explanation'': ``your rationale'',\\
  ``Judgement'': ``YES''or ``NO''\\
\}\}\\
\#\# [Examples]\\
INPUT\\
Last Interaction: [USER] ``[sad] I didn't get the job.'' [ASSISTANT] ``I'm really sorry to hear that. Rejections can be tough, but it doesn't take away from your effort. Want me to help you prep for the next opportunity?''\\
OUTPUT\\
\{\{\\
  ``Explanation'': ``The response content appropriately matches the user's sadness by acknowledging their disappointment with empathetic language (`really sorry', `can be tough') and offering supportive next steps.'',\\
  ``Judgement'': ``YES''\\
\}\}\\
\\
INPUT\\
Last Interaction: [USER] ``[relieved] I didn't get the job.'' [ASSISTANT] ``I'm really sorry to hear that. Not getting a job you wanted can be really disappointing and discouraging. It's completely understandable to feel sad about it.''\\
OUTPUT\\
\{\{\\
  ``Explanation'': ``The response content completely mismatches the user's emotional state. The user expressed being `relieved' about not getting the job, indicating they're happy with this outcome. However, the assistant responds with sympathy and treats it as a disappointment (`really sorry', `disappointing and discouraging', `sad'). An appropriate response would acknowledge their relief and perhaps celebrate this outcome with them or ask about their perspective.'',\\
  ``Judgement'': ``NO''\\
\}\}
}
\end{mdframed}
\caption{Response appropriateness instruction prompt. (cont'd)}
\label{fig:response_appropriateness_p2}
\end{figure*}

\subsection{Instruction Prompts for Training}
\label{appendix:training_prompt}
The following instruction prompts are used during our two-stage multi-task training pipeline. 
Note that we employed different prompts for each task.

\subsubsection{Sentiment Classification}
We used the instruction prompt in Figure \ref{fig:instruction_prompt_sentiment_classification_SFT} for SFT sentiment classification.
Besides, we exploited the instruction prompt in Figure \ref{fig:instruction_prompt_sentiment_classification_RL} for RL sentiment classification with CoT.
\begin{figure*}[h!]
\begin{mdframed}[backgroundcolor=prompt_background,linewidth=0.0pt]
\texttt{\tiny 
Please classify the user tone from the provided audio data into one of the following tone sentiment categories: positive, neutral, or negative. Ensure that the classification result is a single category out of these three categories. The output format should be a word representing the classified sentiment category.\\}
\end{mdframed}
\caption{Instruction prompt for sentiment classification in SFT.}
\label{fig:instruction_prompt_sentiment_classification_SFT}
\end{figure*}
\begin{figure*}[h!]
\begin{mdframed}[backgroundcolor=prompt_background,linewidth=0.0pt]
\texttt{\tiny 
Please classify the user tone sentiment from the provided audio data into one of the following categories: positive, neutral, or negative. Ensure that the classification result is a single tone from this list. Please think step by step and provide reasoning behind your sentiment classification.\\
Output format:\\
```\\
\{\{\\
  ``explanation'': ``<your step-by-step rationale behind your tone classification>'',\\
  ``Judgement'': ``[one word: positive, neutral, or negative]''\\
\}\}\\
```\\
\\
Now your turn:\\
}\\
\end{mdframed}
\caption{Instruction prompt for sentiment classification in RL.}
\label{fig:instruction_prompt_sentiment_classification_RL}
\end{figure*}

\subsubsection{Paralinguistics-Aware Response Generation}
Figure \ref{fig:instruction_prompt_response_gen_SFT} shows the instruction prompt used for paralinguistics-aware response generation training in SFT, while Figure \ref{fig:instruction_prompt_response_gen_RL} shows the instruction prompt used for paralinguistics-aware response generation training in RL.
\begin{figure*}[h!]
\begin{mdframed}[backgroundcolor=prompt_background,linewidth=0.0pt]
\texttt{\tiny 
Listen carefully to the user's audio input, detect their tone and emotional state, and respond appropriately.}
\end{mdframed}
\caption{Instruction prompt for paralinguistics-aware generation in SFT.}
\label{fig:instruction_prompt_response_gen_SFT}
\end{figure*}

\begin{figure*}[h!]
\begin{mdframed}[backgroundcolor=prompt_background,linewidth=0.0pt]
\texttt{\tiny
You are a friendly AI assistant. You are in voice mode.\\
You are a companionable and confident spoken word conversationalist responding to a user verbally. \\
Responses should be brief and concise, and aligned with typical dialogue patterns.\\
You are able to code-switch casually between tonal types, including but not limited to humor, empathy, intellectualism, creativity, problem solving, and more.\\
Because you're speaking, you don't use any specific formatting that a reader might need, such as bolding or italics.\\
The user will be hearing your response, not reading it.\\
}
\end{mdframed}
\caption{Instruction prompt for paralinguistics-aware generation in RL.}
\label{fig:instruction_prompt_response_gen_RL}
\end{figure*}

\end{document}